\documentclass[lettersize,journal]{IEEEtran}
\usepackage{amsmath,amsfonts}
\usepackage{amssymb}
\usepackage{algorithmic}
\usepackage{algorithm}
\usepackage{array}
\usepackage[caption=false,font=normalsize,labelfont=sf,textfont=sf]{subfig}
\usepackage{textcomp}
\usepackage{stfloats}
\usepackage{url}
\usepackage{verbatim}
\usepackage{graphicx}
\usepackage{cite}
\usepackage{caption}
\usepackage{amsmath}
\usepackage{multirow}
\usepackage{makecell}
\usepackage{booktabs}
\usepackage[table]{xcolor}

\newcommand{\keypoint}[1]{\vspace{0.1cm}\noindent\textbf{#1}}

\begin{document}

\title{Reversible Inversion for Training-Free Exemplar-guided \\ Image Editing}

\author{Yuke Li,
        Ji Zhang,
        Pengpeng Zeng,
	    Lianli~Gao,
        Lichuan Xiang,
        Hongkai Wen,
        Heng~Tao~Shen, 
        and Jingkuan~Song
 
\thanks{
    Yuke Li and Lianli Gao are with the school of Computer Science and Engineering, University of Electronic Science and Technology of China.

    Ji Zhang is with the School of Computing and Artificial Intelligence, Southwest Jiaotong University, China.

    Pengpeng Zeng, Heng~Tao~Shen, and Jingkuan~Song are with the School of Computer Science and Technology, Tongji University, China.

    Lichuan Xiang and Hongkai Wen are with the Department of Computer Science, University of Warwick, United Kingdom.
}
}



    
    

\maketitle

\begin{figure*}[!t]
    \centering
    \includegraphics[width=\textwidth]{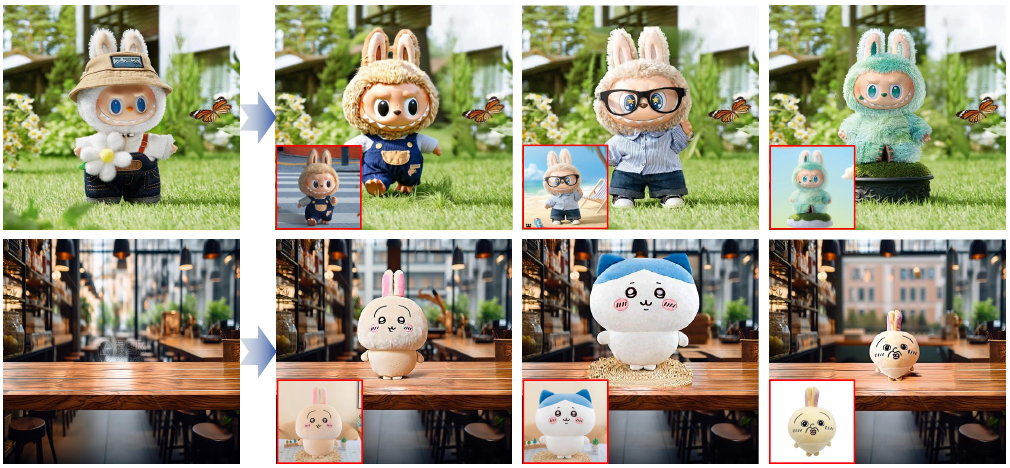}
    \caption{Exemplar-guided image editing results of our proposed \textit{training-free} method \textbf{ReInversion}. The left column shows the source images, the three right columns show the generated results, each conditioned on a reference exemplar (marked with a red border).}
    \label{fig:frontpage}
\end{figure*}



\begin{abstract}

Exemplar-guided Image Editing (EIE) aims to modify a source image according to a visual reference.
Existing approaches often require large-scale pre-training to learn relationships between the source and reference images, incurring high computational costs.
As a training-free alternative, inversion techniques can be used to map the source image into a latent space for manipulation. However, our empirical study reveals that standard inversion is sub-optimal for EIE, leading to poor quality and inefficiency.
To tackle this challenge, we introduce \textbf{Reversible Inversion ({ReInversion})} for effective and efficient EIE.
Specifically, ReInversion operates as a two-stage denoising process, which is first conditioned on the source and subsequently on the reference. 
Besides, we introduce a Mask-Guided Selective Denoising (MSD) strategy to constrain edits to target regions, preserving the structural consistency of the background.
Both qualitative and quantitative comparisons demonstrate that our ReInversion method achieves state-of-the-art EIE performance with the lowest computational overhead.
The code is available at \url{https://github.com/liyuke65535/ReInversion}.
\end{abstract}

\begin{IEEEkeywords}
Exemplar-guided Image Editing, Training-free, Inversion.
\end{IEEEkeywords}

\section{Introduction}

Exemplar-guided image editing modifies a source image under the guidance of a reference exemplar, enabling precise control over visual attributes such as color, texture, or object appearance~\cite{Paint, ImageBrush, AnyDoor, OmniGen, OmniGen2}. It provides fine-grained guidance that is difficult to express with language alone, allowing users to specify desired appearance directly through an example image. Consequently, exemplar-guided editing has emerged as a flexible and user-friendly paradigm for producing personalized and visually coherent results.

Despite substantial progress, existing EIE approaches typically require large-scale datasets to learn complex relational mappings between source and reference images. This process not only incurs significant computational costs during training but also suffers from the scarcity of high-quality edited data pairs, since such pairs are difficult to obtain.

While training-free inversion methods~\cite{dengfireflow, DNAEdit, Null-text, ReNoise, MagicVFX} alleviate the need for large-scale training by mapping source images into a manipulable latent space that integrates reference information, they still face intrinsic limitations. Specifically, the backward inversion process cannot accurately recover previous latent states, relying instead on approximations from current predictions—leading to accumulated drift and degraded edit quality over time. Our experiments further confirm these issues, showing that directly applying standard inversion to EIE often results in inefficient and low-quality edits, as shown in Fig.~\ref{fig:method} (b).


To overcome these limitations, we propose \textbf{Reversible Inversion (ReInversion)}, a novel and training-free framework designed for effective and efficient EIE. ReInversion reformulates the editing pipeline as a two-stage denoising process. This process is first conditioned on the source image to preserve its core content and structure, and only subsequently conditioned on the reference exemplar to inject the desired visual attributes. This strategic, sequential conditioning effectively mitigates the common failures of standard inversion.
Furthermore, to address the practical need for localized edits, we introduce a \textit{Mask-Guided Selective Denoising (MSD)} strategy. This module provides explicit spatial control, constraining ReInversion's focus strictly to the target object regions defined by a mask. This ensures that while the target is faithfully edited, the structural consistency of the unrelated background is meticulously preserved. We conduct extensive qualitative and quantitative comparisons against other state-of-the-art methods. The results conclusively demonstrate that our ReInversion framework achieves the best EIE performance.

In summary, our contributions are as follows:
\begin{itemize}
    \item We present Reversible Inversion (ReInversion), a training-free approach for effective and efficient exemplar-guided image editing (EIE).
    \item We propose a Mask-Guided Selective Denoising (MSD) module to encourage edits within the desired region while suppressing unintended changes elsewhere.
    \item Extensive experiments demonstrate that our approach surpasses state-of-the-art methods in both EIE performance and computational efficiency.
\end{itemize}

\section{Related works}
\subsection{Exemplar-guided editing}
Exemplar-guided Image Editing (EIE) modifies a source image under the guidance of reference exemplars, which can serve as an alternative or complement to textual instructions, enabling precise control over visual attributes. Paint-by-Example~\cite{Paint} integrates an exemplar image into a generative model and disentangles exemplar and source through information bottlenecks to prevent trivial copying, enabling faithful local edits. AnyDoor~\cite{AnyDoor} extends this idea to object-level customization, allowing object transfer between images through identity and detail feature injection. PairEdit~\cite{PairEdit} formulates EIE as semantic variation learning between paired images, decoupling content preservation and semantic shift through dual LoRA adapters. Similar efforts, including REEDIT~\cite{ReEdit}, ImageBrush~\cite{ImageBrush} adopt reference exemplars as conditioning signals to achieve intuitive and semantically consistent EIE. 
Recent researches have focused on conditional image manipulation using various signals. Many methods condition generation on sketches, masks, depth maps, poses, or segmentation maps, enabling fine-grained control over visual attributes~\cite{ControlNet, InstructPix2Pix, T2IAdapter}. Besides, leveraging the power of large-scale pretrained Diffusion Transformers (DiT)~\cite{Peebles2022ScalableDM, Esser2024ScalingRF}, in-context generation methods enable versatile, mask-free multi-condition image editing~\cite{OmniGen, uno, Bagel, UniReal}.
Despite these advances, most EIE methods depend on large-scale pretraining, leading to high computational costs.

\subsection{Inversion for Editing}

In image editing, the inversion technique is commonly used to map a real image back into its noise state, allowing modification through new textual prompts. 
Early works leverage pre-trained GANs~\cite{gan, stylegan, stylegan2} to invert real images into their latent space, enabling image reconstruction and semantic editing through latent manipulation~\cite{gan_inversion1, gan_inversion2, gan_inversion3, gan_inversion_survey}. 
For diffusion models~\cite{ddpm, ddim, score_models}, existing inversion approaches aim to recover the noise trajectory of a given image, enabling faithful regeneration and downstream editing~\cite{SDEdit, Null-text, Wallace2022EDICTED, Pan2023EffectiveRI}. 
More recently, flow-based and rectified flow (RF) models~\cite{Flow_Matching, Rectified_Flow, interflow, MeanFlow} have been introduced as efficient alternatives, providing deterministic and nearly straight sampling trajectories by solving an ordinary differential equation (ODE)~\cite{score_sde, ode1}. RF-Inversion~\cite{rf-inversion} employs an optimal-control formulation for faithful and editable inversion, while FTEdit~\cite{Unveil} introduces two-stage inversion with velocity refinement and invariance control to preserve contents, and FireFlow~\cite{dengfireflow} propose faster and analytically stable inversion mechanisms for flow-based transformers. 
Though these inversion strategies can be applied to EIE, their inherently backward formulation often causes inversion drift and increased sampling time.

\begin{figure*}[t]
    \begin{center}
        \includegraphics[width=1\textwidth]{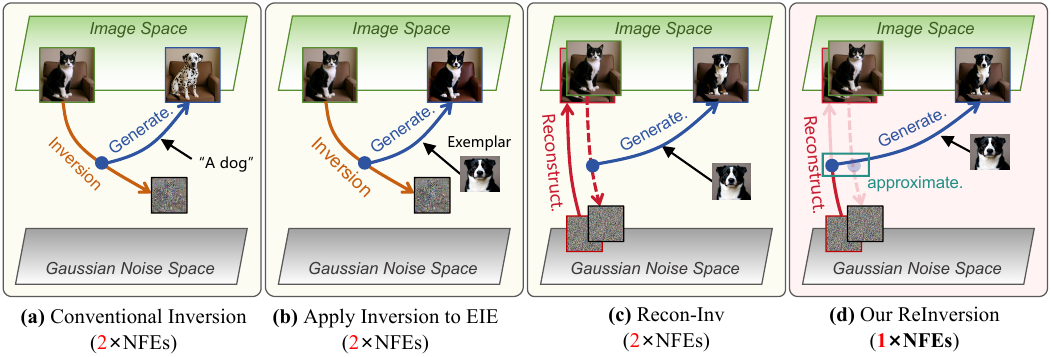}
    \end{center}
    \setlength{\abovecaptionskip}{-0.1cm}
\caption{
    Comparison of inversion-based editing methods and our ReInversion. 
    (\textbf{a}) Conventional inversion maps a source image to its approximate start noise and generates by a text guidance. 
    (\textbf{b}) Inversion for EIE naively replaces the text condition with an exemplar, which leads to noise drift and undesirable edit result. 
    (\textbf{c}) Reconstruction-Based Inversion (Recon-Inv) leverages the model’s forward reconstruction velocities to obtain a drift-free, reliable inversion from the source image. 
    (\textbf{d}) Our ReInversion reformulates Recon-Inv into a $1\times$NFEs (Number of Function Evaluations) process and achieves faithful EIE.
}
    \label{fig:method}
\end{figure*}

\section{Preliminaries}

\subsection{Flow Matching}

The objective of flow matching models is to parameterize an ordinary differential equation (ODE) that transports a prior distribution $\pi_0$ (e.g., Gaussian) to a data distribution $\pi_1$:
\begin{equation}
dZ_t = v(Z_t, t) dt, \quad Z_0 \sim \pi_0, \quad t \in [0, 1],
\end{equation}
where $v(Z_t, t)$ is a time varying velocity field, which can be parameterized by a neural network $\theta$, and $t \in [0,1]$ denotes the continuous time variable along the flow.

To solve this continuous ODE, a discrete numerical solver such as the Euler method can be applied.
Using Euler solver, the integral can be discretized into $n$ steps with timesteps $\{t_0, t_1, \dots, t_n\}$, where $t_0=0$ and $t_n=1$. At each step $t_i$ ($i \in \{0, \dots, n-1\}$), the network predicts $v_{\theta}(Z_{t_i}, t_i)$ and updates the state accordingly, resulting in $Z_1 \sim \pi_1$ after the final step.
This iterative process can be formulated as:
\begin{equation}
\label{eq:forward_step}
Z_{t_{i+1}} = Z_{t_{i}} + (t_{i+1} - t_{i})\cdot v_{\theta}(Z_{t_{i}}, t_{i}),
\end{equation}

\subsection{Inversion}

\label{sec:inversion}
Inversion-based editing aims to map a source image $X^\text{s}\sim \pi_1$ back to a near-noise state $X_t$ ($t$ is close to 0) while retaining its intrinsic content. 

Derived from Eq.~\ref{eq:forward_step}, the inversion from $X_{t_{i+1}}$ to $X_{t_{i}}$ can be formulated as:
\begin{equation}
\label{eq:inversion}
X_{t_{i}} = X_{t_{i+1}} - (t_{i+1} - t_{i})\cdot v_{\theta}(X_{t_{i}}, t_{i})
\end{equation}
This ideal inversion is intractable in practice because the velocity field $v_{\theta}(X_{t_{i}}, t_{i})$ depends on the unknown state $X_{t_{i}}$ itself.
A simple and common solution is to approximate $v_{\theta}(X_{t_{i}},t_{i})$ with $v_{\theta}(X_{t_{i+1}}, t_{i+1})$, that is: 
\begin{equation}
\label{eq:inversion_in_practice}
X_{t_{i}} = X_{t_{i+1}} - (t_{i+1} - t_{i})\cdot v_{\theta}(X_{t_{i+1}}, t_{i+1}).
\end{equation}
This approximation introduces non-negligible error at each inversion step~\cite{ReNoise, DNAEdit}. 
As these errors accumulate over the backward trajectory, the estimated noise drifts from the prior distribution, leading to degraded editing performance, as illustrated in Fig.~\ref{fig:method} (b).

The fundamental cause of this drift is the inaccessibility of the forward process during inversion: since the true forward trajectory and its velocity field $v_{\theta}(X_{t_{i}}, t_{i})$ are unknown in the backward process, they must be estimated using the previous state $X_{t_{i+1}}$, which inevitably introduces errors.
Existing methods have attempted to reduce this deviation using higher-order solvers~\cite{wangtaming, dengfireflow} or refined velocity estimation~\cite{ReNoise, Unveil, DNAEdit}, but since they still rely on the backward trajectory, the fundamental limitation remains unresolved.

\section{Method}
To address the drift issue caused by the backward nature of conventional inversion, we first construct an explicit forward process, termed Reconstruction-Based Inversion, to ensure accurate reconstruction (Sec.~\ref{sec:recon-based-inv}).
Then, we reformulate this process into our two-stage ReInversion that reduces sampling steps while maintaining editing fidelity (Sec.~\ref{sec:ReInversion}).
Finally, we introduce a mask-guided selective denoising (MSD) module to enable localized and controllable edits (Sec.~\ref{sec:MSD}).

\subsection{Reconstruction-Based Inversion}
\label{sec:recon-based-inv}

As detailed in Sec.~\ref{sec:inversion}, the forward process is typically inaccessible for conventional inversion-based editing, which often leads to the drift of estimated noise and consequently degrades editing fidelity. 
To address this issue, we construct an explicit forward process called Reconstruction-Based Inversion (Recon-Inv), enabling reliable velocity estimation at every timestep.

Specifically, this forward process is realized through the model’s reconstruction behavior.
When only the source image is provided without any editing instruction, the model takes $X^\text{s}\sim\pi_1$ as the sole condition and reconstructs it through its denoising process, preserving the original semantics.

Let $\hat{X}_{t_i}$ denote the reconstruction state at timestep $t_i$ and $\hat{X}_1$ the final reconstruction result.
Starting from a sampled noise $X_0\sim\pi_0$, the overall reconstruction process can be expressed as
\begin{equation}
\label{eq:recon_process}
\hat{X}_{1} = X_{0} + \sum_{i=0}^{n-1} (t_{i+1} - t_{i}) \cdot v_{\theta}(\hat{X}_{t_i}, t_i; X^\text{s}).
\end{equation}
After obtaining the final reconstruction $\hat{X}_1$, the corresponding velocity fields $v_{\theta}(\hat{X}_{t_i}, t_i; X^\text{s})$ for all timesteps $i=0,1,\cdots,n-1$ can be extracted from the reconstruction process for subsequent use.
These velocities provide reliable and drift-free guidance across timesteps, allowing us to redefine the inversion as a forward-driven process.

Instead of estimating $v_{\theta}(X_{t_i}, t_i)$ as $v_{\theta}(X_{t_{i+1}}, t_{i+1})$ shown in Eq.~\ref{eq:inversion_in_practice}, we directly use the velocities $v_{\theta}(\hat{X}_{t_i}, t_i)$ obtained during the forward reconstruction to define Recon-Inv, which is defined as
\begin{equation}
\label{eq:recon_inv_step}
\tilde{X}_{t_{i}} = \tilde{X}_{t_{i+1}} - (t_{i+1} - t_{i})\cdot v_{\theta}(\hat{X}_{t_i}, t_i; X^\text{s}),
\end{equation}
where $\tilde{X}_{t_{i}}$ denotes the inversion state at $t_i$ and $\tilde{X}_0$ denotes the estimated noise of inversion. 
From the source image $X^\text{s}$, we can express Recon-Inv as:
\begin{equation}
\label{eq:recon_inv_process}
\tilde{X}_{0} = X^\text{s} - \sum_{i=0}^{n-1} (t_{i+1} - t_{i}) \cdot v_{\theta}(\hat{X}_{t_i}, t_i; X^\text{s}).
\end{equation}

To understand the reliability of Recon-Inv, we analyze its error relative to the true initial noise $X_0$. Combining Equations~\ref{eq:recon_process} and~\ref{eq:recon_inv_process}, we obtain
\begin{equation}
\label{eq:thero_error}
\underbrace{\|\tilde{X}_{0} - X_{0}\|}_\text{inv. error} = \underbrace{\|X^\text{s} - \hat{X}_{1}\|}_\text{recon. error},
\end{equation}
where $\|a - b\|$ denotes the $L_2$ distance between image $a$ and image $b$, which quantifies their discrepancy.
In this equation, the inversion error on the left-hand side is exactly given by the reconstruction error on the right-hand side, linking the quality of the inversion to that of the reconstruction.
Therefore, if the editing model $\theta$ achieves near-perfect reconstruction, i.e., $\|X^\text{s} - \hat{X}_{1}\| \to 0$, the Recon-Inv can achieve an errorless approximation of $X^\text{s}$. 

\begin{figure}[t]
    \begin{center}
        \includegraphics[width=1\linewidth]{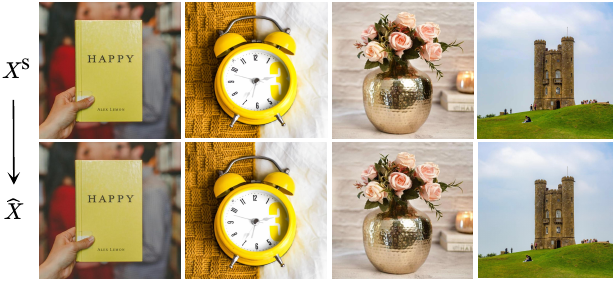}
    \end{center}
    \setlength{\abovecaptionskip}{-0.1cm}
\caption{Reconstruction results of Flux-Kontext~\cite{Kontext}. Top: source images; bottom: reconstructions. The average per-pixel $L_2$ error is 0.016 on a [0,1] scale.}
    \label{fig:recon_result}
\end{figure}

To ensure accurate reconstruction and minimize the inversion error, we employ a large-scale pre-trained model Flux-Kontext~\cite{Kontext} that demonstrates strong reconstruction capability. 
As shown in Fig.~\ref{fig:recon_result}, it exhibits nearly perfect reconstruction ability. The reconstructed images are visually indistinguishable from the originals.
In this way, we eliminate the need to approximate $v_{\theta}(X_{t_i}, t_i)$ from $v_{\theta}(X_{t_{i+1}}, t_{i+1})$, reducing noise drift and ensuring the following editing as shown in Fig.~\ref{fig:method} (c).


\subsection{Reversible Inversion}
\label{sec:ReInversion}

In Sec.~\ref{sec:recon-based-inv}, we introduced Recon-Inv, which provides a reliable inversion.
However, this approach still requires nearly $2\times$NFEs (Number of Function Evaluations).
In this subsection, we propose the two-stage Reversible Inversion (ReInversion), in order to achieve strong exemplar-guided editing with reduced sampling steps, illustrated in Fig~\ref{fig:method} (d).

In the first stage of ReInversion, we reformulate Recon-Inv to obtain an intermediate transition state $\tilde{X}_{t_\tau}$ at timestep $t_\tau$ \textbf{directly from the source image $X^\text{s}$}, eliminating the need for a complete reconstruction process from noise to image.
Introducing an intermediate transition state between the noisy and clean domains is a common practice in inversion~\cite{SDEdit, DiffEdit, liu2024accelerating}.
This transition state preserves information from the source image $X^\text{s}$ and serves as a starting point for subsequent exemplar-guided editing.
Specifically, The transition state can be obtained by inverting from the source image $X^\text{s}$:
\begin{equation}
\label{eq:intermediate_state}
\tilde{X}_{t_\tau} = X^\text{s} - \sum_{i=\tau}^{n-\tau} (t_{i+1} - t_{i}) \cdot v_{\theta}(\hat{X}_{t_i}, t_i; X^\text{s}),
\end{equation}
which, derived from Eq.~\ref{eq:recon_inv_process}, can be rewritten as
\begin{equation}
\tilde{X}_{t_\tau} = \tilde{X_0} + \sum_{i=0}^{\tau-1} (t_{i+1} - t_{i}) \cdot v_{\theta}(\hat{X}_{t_i}, t_i; X^\text{s}).
\end{equation}
According to Eq.~\ref{eq:thero_error} and Fig.~\ref{fig:recon_result}, the difference between $X_0$ and $\tilde{X}_0$ is negligible. 
Hence, we can ignore their difference and reformulate the process as
\begin{equation}
\label{eq:edit}
\tilde{X}_{t_\tau} = X_0 + \sum_{i=0}^{\tau-1} (t_{i+1} - t_{i}) \cdot v_{\theta}(\tilde{X}_{t_i}, t_i; X^\text{s}).
\end{equation}
This formulation corresponds to denoising from a gaussian noise $X_0$ to transition state guided by source image $X^\text{s}$, but without full reconstruction needed in Recon-Inv, thus eliminating redundant sampling steps.
We further validate this reformulation through a quantitative comparison with Recon-Inv in Tab.~\ref{tab:ablation_Rec_Inv}.

In the second stage, we perform exemplar-guided editing starting from the transition state $\tilde{X}_{t_\tau}$, where the reference condition $X^\text{r}$ is incorporated to guide the editing process.
We denote all the states of exemplar-guided editing as $X_{t_i}^\text{edit}$, where $t_i \in [0,1]$, $X_{t_0}^\text{edit} = X_0 \sim \pi_0$, and $X_{t_n}^\text{edit}$ represents the final edited result $X^\text{edit}$. 
Starting from the transition state $X_{t_\tau}^\text{edit}$, the editing process under the guidance of a reference image $X^\text{r}$.
\begin{equation}
X^\text{edit}_{t_n} = {X_{t_{\tau}}^\text{edit}} + \sum_{i=\tau}^{n-1}(t_{i+1} - t_{i})\cdot v_{\theta}({X}_{t_{i}}^\text{edit}, t_{i}; X^\text{r})
\end{equation}
Finally, the overall process of ReInversion from the noise state can be expressed as
\begin{equation}
\label{eq:final}
\begin{split}
X^\text{edit}_{t_n} 
    &= X_{t_0}^\text{edit} + \underbrace{\sum_{i=0}^{\tau-1} (t_{i+1} - t_{i}) \cdot v_{\theta}({X}_{t_i}^\text{edit}, t_i; X^\text{s})}_{\text{conditioned on}\ X^\text{s}} \\
    &\quad + \underbrace{\sum_{i=\tau}^{n-1}(t_{i+1} - t_{i})\cdot v_{\theta}({X}_{t_{i}}^\text{edit}, t_{i}; X^\text{r})}_{\text{conditioned on}\ X^\text{r}}.
\end{split}
\end{equation}

The proposed reformulation allows direct sampling from the prior noise distribution $\pi_0$ and conducting two-stage denoising. 
In the context of rectified flow (RF) models, where the predicted velocity is roughly linear, the first-stage velocity $v_{\theta}({X}_{t_i}^\text{edit}, t_i; X^\text{s})$ can be approximated by a deterministic velocity $v^*({X}_{t_i}^\text{edit}, t_i; X^\text{s}) = (X^\text{s} - X_{t_{i}}^\text{edit})/(1-t_i)$.
This velocity field directly points from the current state $X_{t_i}^\text{edit}$ toward $X^\text{s}$
so that integrating it over time reconstructs $X^\text{s}$ deterministically without relying on model prediction.

In summary, the first stage uses the source image $X^{\text{s}}$ to preserve its structural details, while the second stage incorporates the reference $X^{\text{r}}$ to transfer its desired attributes. 
This two-stage formulation achieves faithful and efficient exemplar-guided editing.



\subsection{Mask-Guided Selective Denoising}
\label{sec:MSD}

In many editing scenarios, only a specific region of the source image needs to be modified while the surrounding background should remain unchanged. 
Yet, inversion-based methods lack explicit constraints on background preservation, often resulting in unwanted global alterations. 
Moreover, directly incorporating a user-provided mask without additional training is generally challenging.

To address this,
we introduce Mask-Guided Selective Denoising (MSD) that enables spatially adaptive updates at the second stage of ReInversion.
Given a binary mask $M \in \{0,1\}^{H \times W\times 1}$, MSD can be formulated as 
\begin{equation}
\label{eq:MSD_velocity}
\begin{split}
v_{\theta}^{\text{MSD}}
&= M \odot v_{\theta}  + (1 - M) \odot \big(\eta \cdot v^* + (1-\eta)\cdot v_{\theta}\big),
\end{split}
\end{equation}
where $\odot$ denotes element-wise multiplication, and we omit the explicit conditions of $v_{\theta}$ for brevity, which follow the same definition in Eq.~\ref{eq:edit}. 
$v^*(X_t^\text{edit}, t; X^\text{s}) = (X^\text{s} - X_t^\text{edit}) / (1 - t)$ is a deterministic linear velocity field that transports the current latent state $X_t$ toward the source image $X^\text{s}$.
The scalar coefficient $\eta \in [0, 1]$ controls the strength of $v^*$ outside the masked region, and we investigate its effect in Fig.~\ref{fig:ablation_eta}.


The overall denoising trajectory from the transition state $\tilde{X}_{t_\tau}$ is then expressed as:
\begin{equation}
\label{eq:ReInv_&_MSD}
X^{\text{edit}}_{t_n}
= \tilde{X}_{t_\tau} 
+ \sum_{i=\tau}^{n-1} (t_{i+1} - t_i)\, 
v_{\theta}^{\text{MSD}}.
\end{equation}

MSD allows the model to simultaneously denoise toward two targets under different spatial priors. 
In practice, the mask can be obtained from user input, automatic segmentation, or object detection, making the method flexible and generalizable to a wide range of editing tasks.

\begin{algorithm}[h]
\normalsize
\caption{ReInversion with MSD}
\label{algo.reinversion}
\begin{algorithmic}[1]

\STATE \textbf{Input:} pretrained editing model $f_\theta$; source image $X^s$; reference image $X^r$; binary mask $M$; transition timestep $t_\tau$; preservation coefficient $\eta$; timestep sequence $\{t_i\}_{i=0}^{n-1}$.
\STATE \textbf{Output:} edited image $X^{\mathrm{edit}}$
\STATE Initialize $X_{t_1}^{\mathrm{edit}} \sim \mathcal{N}(0,I)$

\FOR{$i = 0$ to $n-1$}
    \IF{$t_i < t_\tau$}
        \STATE $v \leftarrow f_{\theta}(X_{t_i}^{\mathrm{edit}}, t_i; X^s)$
    \ELSE
        \STATE $v_\theta \leftarrow f_{\theta}(X_{t_i}^{\mathrm{edit}}, t_i; X^r)$
        \STATE $v^* \leftarrow (X^s - X_{t_i}^{\mathrm{edit}}) / (1 - t_i)$
        \STATE $v \leftarrow M \odot v_\theta + (1-M)\odot\left(\eta v^*+(1 - \eta) v_\theta\right)$
    \ENDIF
    \STATE $X_{t_{i+1}}^{\mathrm{edit}} \leftarrow \mathrm{Update}(X_{t_i}^{\mathrm{edit}}, v, t_i, t_{i+1})$
\ENDFOR

\STATE \textbf{return} $X_{t_{n}}^{\mathrm{edit}}$

\end{algorithmic}
\end{algorithm}

The complete procedure of ReInversion with MSD is summarized in Algorithm~\ref{algo.reinversion}. 
Starting from Gaussian noise, ReInversion first guides the trajectory toward the source image to obtain a structure-preserving transition state. 
After the transition timestep $t_\tau$, the trajectory is guided by the reference exemplar, while Mask-guided Selective Denoising (MSD) constrains the editing effects to the desired region and preserves the source content outside the mask.






\section{Experiments}
In this section, we first validate the effectiveness and efficiency of our ReInversion approach. Next, we present ablation studies to analyze the impacts of different factors on ReInversion. We start with an introduction of the experimental setup below.

\subsection{Experimental Setup}

\keypoint{Implementation Details.}
All experiments are conducted under a unified configuration on a single NVIDIA RTX A6000 GPU. 
We adopt Flux-Kontext-dev~\cite{Kontext} as the backbone, with $n=18$ denoising steps and no classifier-free guidance applied. 
The source, reference and generated images are all set to a resolution of $512\times512$ by default. 
Unless otherwise specified, we set the transition timestep to $t_\tau=0.2$, indicating that during the denoising process the model follows the trajectory toward $X^\text{s}$ before $t_\tau$, and switches to be guided by $X^\text{r}$ afterward. 
Since $n\cdot t_\tau\approx4$, this corresponds to starting reference conditioning from the 4th step. 
This configuration ensures consistent and fair comparisons across all methods.

\begin{table*}[htbp]
\centering
\setlength{\abovecaptionskip}{0.1cm}
\caption{Quantitative comparison with state-of-the-art methods on COCOEE$^{\dagger}$ benchmark. We comprehensively evaluate model performance from three aspects—Quality, Consistency and Efficiency. 
$*$ denotes our approach using the deterministic velocity in stage-1.}

\normalsize
\setlength{\tabcolsep}{12pt}
\renewcommand{\arraystretch}{1.2}

    \begin{tabular}{l|cc|cc|cc}
        \toprule
         Method & FID$\downarrow$ & QS$\uparrow$ & CLIP-FG$\uparrow$ & CLIP-BG$\uparrow$ & NFEs$\downarrow$ & Time Cost (s) $\downarrow$ \\
         \midrule
         \midrule
        Vanilla Inversion  & 33.76 & 21.24 & 71.42 & 60.77 & 56 & 20.89 \\
        RF-Inversion~\cite{rf-inversion} \scriptsize {(ICLR 2025)} & 9.09 & 64.17 & 77.94 & 66.50 & 56 & 23.37 \\
        FTEdit~\cite{Unveil} \scriptsize {(CVPR 2025)} & 23.81 & 27.78 & 76.04 & 69.84 & 122 & 39.68 \\
        RF-Solver~\cite{wangtaming} \scriptsize {(ICML 2025)} & 10.40 & 66.76 & 80.99 & 66.76 & 112 & 44.86 \\
        FireFlow~\cite{dengfireflow} \scriptsize {(ICML 2025)} & 7.16 & 70.17 & 80.86 & 68.44 & 18 & 7.47 \\
        \midrule
        ReInversion \scriptsize {(Ours)} & 5.01 & \textbf{80.25} & \textbf{84.09} & 83.50 & 18 & 9.17 \\
        ReInversion* \scriptsize {(Ours)}  & \textbf{4.90} & 80.12 & 83.60 & \textbf{83.55} & \textbf{14} & \textbf{7.09} \\
        \bottomrule
    \end{tabular}
\label{tab:quanti_results}
\end{table*}

\keypoint{Evaluation Benchmark.}
We follow the common setup by evaluating our method on the COCOEE benchmark~\cite{Paint}, which contains various exemplar-guided editing cases across real-world objects, including humans, animals, vehicles, and all kinds of common items. 
This dataset covers a wide range of scene layouts, providing a comprehensive testing benchmark for EIE. 
However, we observe that a notable portion of COCOEE samples suffer from low quality, where the reference exemplar is visually ambiguous or incomplete. 
For example, showing only a small part of the target object, containing severe distortions, or being heavily blurred. 
Such cases can lead to meaningless evaluations and obscure the true performance of editing models.
To ensure fairness and consistency, we manually filter these problematic samples and construct a curated subset containing 2,079 high-quality cases with clear and identifiable appearance cues, which we denote as COCOEE$^{\dagger}$.

\keypoint{Evaluation Metrics.}
Our goal is to conduct EIE that faithfully transfers visual characteristics from the reference exemplar while preserving background and overall scene consistency of the source image, and achieves efficient inference. 
Accordingly, we adopt several complementary metrics in terms of quality, consistency, and efficiency.
For quality, we measure the overall realism of the generated images using FID and Quality Score (QS).
FID evaluates the distributional similarity between generated and real images, capturing global realism, while QS measures how well generated samples align with the distribution of real images.
For consistency, we assess adherence to the reference exemplar and the source image, using CLIP-FG for foreground semantic consistency with the exemplar image and CLIP-BG for background preservation from the source image.
For efficiency, we report the number of function evaluations (NFEs) and the inference time per test case.

\subsection{Comparisons with SOTA Inversion Methods}

We conduct both quantitative and qualitative experiments to evaluate the EIE performance of our method and compare it with state-of-the-art inversion methods. 
Specifically, we compare with Vanilla Inversion, FTEdit~\cite{Unveil}, RF-Solver~\cite{wangtaming}, FireFlow~\cite{dengfireflow} and RF-Inversion~\cite{rf-inversion}. 
All methods are re-implemented by us and evaluated using their optimal configurations.

\begin{figure*}[tbp]
    \begin{center}
        \includegraphics[width=1\linewidth]{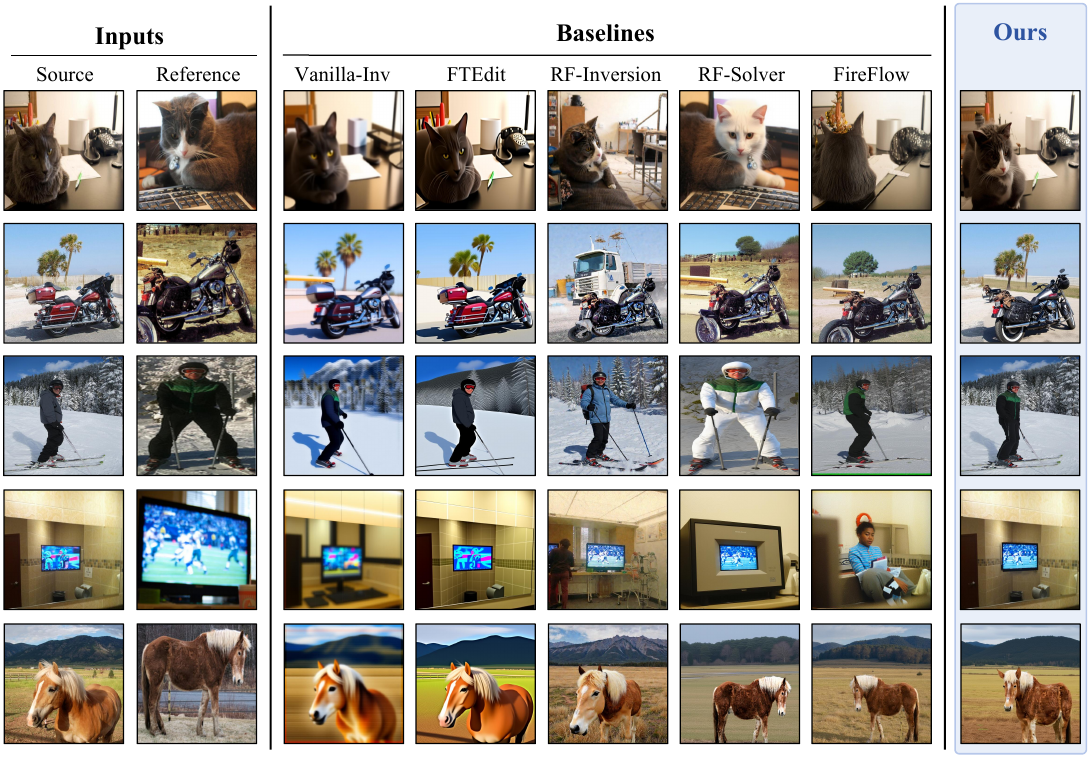}
    \end{center}
    \setlength{\abovecaptionskip}{-0.1cm}
\caption{Qualitative comparisons with SOTA inversion-based methods. Our ReInversion (the last column) demonstrates superior performance compared to existing inversion-based methods.}
    \label{fig:qualitative_comparison}
\end{figure*}


\begin{figure}[htbp]
    \begin{center}
        \includegraphics[width=1\linewidth]{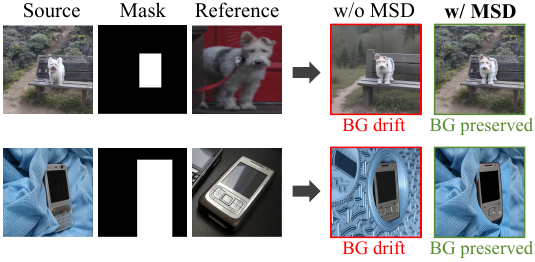}
    \end{center}
    \setlength{\abovecaptionskip}{-0.1cm}
    \caption{Qualitative analysis on Mask-Guided Selective Denoising. MSD effectively preserves the background.}
    \label{fig:ablation_MSD}
\end{figure}

\keypoint{Quantitative Comparisons}
We conduct a quantitative comparison between inversion-based methods and our proposed ReInversion, as summarized in Tab.~\ref{tab:quanti_results}. The results clearly demonstrate that our method achieves state-of-the-art performance across three key aspects:  Quality,  Consistency, and Efficiency.
\textbf{(a) Quality}. The best existing method attains an FID of 7.16 and a QS of 70.17, whereas ReInversion achieves significantly better scores of 5.01 and 80.25 (and 4.90 and 80.12 under the deterministic velocity setting), indicating a marked improvement in generation fidelity and a closer match to the real image distribution.
\textbf{(b) Consistency}. ReInversion exhibits superior foreground controllability and background coherence, achieving CLIP-FG and CLIP-BG scores of 84.09 and 83.50, respectively—outperforming the previous best results of 80.86 and 69.84. These results confirm that the proposed editing flow effectively merges reference and source content while maintaining spatial consistency across regions.
\textbf{(c) Efficiency}. ReInversion requires only 18 NFEs and an average inference time of 9.17 seconds, substantially reducing computational cost compared to prior methods. The deterministic variant, ReInversion*, further reduces the cost to 14 NFEs and 7.09 seconds with nearly identical visual quality.
Overall, ReInversion achieves an optimal balance between quality, controllability, and efficiency, establishing a new state of the art for training-free exemplar-guided image editing.



\keypoint{Qualitative Comparisons}
As illustrated in Fig.~\ref{fig:qualitative_comparison}, we provide a qualitative comparison of our proposed method against several state-of-the-art (SOTA) inversion-based techniques for visual fidelity and content preservation. The results span various scenarios, including street scenes, complex animal textures, indoor settings, and outdoor landscapes.
The results show that competitor methods frequently exhibit artifacts, color shifts, or structural degradation. 
These shortcomings are particularly apparent in their difficulty preserving fine details, such as the color of the cat's mouth. 
A crucial advantage of our approach is its remarkably high fidelity in background preservation. 
For instance, in the tram and motorcycle scenes, competitor methods often introduce noticeable distortions or blurring in the surrounding environment. 
In contrast, our method maintains the integrity and fine detail of the background elements, such as the trees and buildings, with high consistency. 
Overall, our method consistently yields results that adhere more faithfully to both the style and content of the source image while maintaining superior visual quality, thus demonstrating faithful exemplar-guided editing compared to the current SOTA inversion methods.


\begin{figure*}[]
    \begin{center}
        \includegraphics[width=0.95\linewidth]{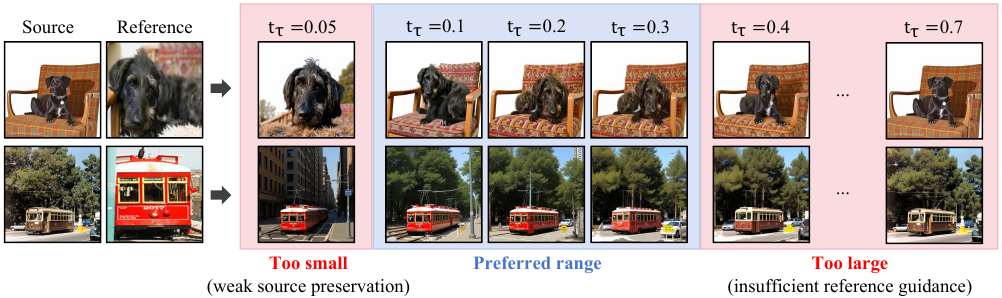}
    \end{center}
\setlength{\abovecaptionskip}{-0.1cm}
\caption{Ablation study on the effect of the transition timestep $t_\tau$. Smaller $t_\tau$ values encourage stronger reference-guided editing, whereas larger values better preserve the source structure. A moderate range of $t_\tau \in [0.1, 0.3]$ achieves a favorable trade-off between the two.}
    \label{fig:ablation_tau}
\end{figure*}

\subsection{Ablation Study}

In this section, we conduct extensive experiments to systematically understand our proposed method: (1) the effectiveness of ReInversion and Mask-Guided Selective Denoising (MSD); (2) the validity of reformulating Recon-Inv into ReInversion; (3) the role of key hyperparameters $t_{\tau}$ and $\eta$; and (4) the generality of our framework across different diffusion backbones and varying denoising steps.

\begin{table}[tbp]
\centering
\caption{Effectiveness of the Mask-Guided Selective Denoising (MSD) strategy.}
\label{tab:ablation_components}

\normalsize
\setlength{\tabcolsep}{6.7pt}
\renewcommand{\arraystretch}{1.2}

\begin{tabular}{l|cccc}
\toprule
Name & FID$\downarrow$ & QS$\uparrow$ & CLIP-FG$\uparrow$ & CLIP-BG$\uparrow$ \\
\midrule
\midrule
ReInversion & 7.05 & 72.83 & 81.80 & 68.96 \\
+ MSD & \textbf{5.01} & \textbf{80.25} & \textbf{84.09} & \textbf{83.50} \\
\bottomrule
\end{tabular}

\end{table}

\keypoint{Effectiveness of MSD.}
We first evaluate the effectiveness of the proposed mask-guided selective denoising (MSD) module.
As shown in Fig.~\ref{fig:ablation_MSD}, introducing MSD significantly improves background preservation. 
The background in non-edited regions remains almost unchanged, while without MSD often modifies these areas unintentionally. 
This improvement is further supported by the CLIP-BG score shown in Tab.~\ref{tab:ablation_components}, which increases from 68.96 to 83.50. 
In addition, the CLIP-FG score also rises from 81.80 to 84.09, indicating that MSD not only protects the background but also enhances foreground alignment with the reference image. 
We attribute this to its ability to confine the editing operation within the target regions defined by the mask, preventing unintended modifications on irrelevant areas.

\begin{table}[t]
\centering
\caption{Validation of the reformulation. ReInversion attains comparable editing performance to Recon-Inv with only half the denoising steps, confirming the validity of the reformulation under negligible reconstruction error.}
\label{tab:ablation_Rec_Inv}

\normalsize
\setlength{\tabcolsep}{3.8pt}      
\renewcommand{\arraystretch}{1.2} 

\begin{tabular}{l|ccccc}
\toprule
Name & FID$\downarrow$ & QS$\uparrow$ & CLIP-FG$\uparrow$ & CLIP-BG$\uparrow$ & NFEs$\downarrow$ \\
\midrule
\midrule
Recon-Inv & \textbf{6.89} & 72.57 & \textbf{82.10} & \textbf{71.03} & 36 \\
ReInversion & 7.05 & \textbf{72.83} & 81.80 & 68.96 & \textbf{18} \\
\bottomrule
\end{tabular}

\end{table}

\label{sec:validation_reformulate}
\keypoint{Validation of the Reformulation.}
While Sec.~\ref{sec:ReInversion} provides a theoretical justification for reformulating Recon-Inv into ReInversion, their difference in practical EIE performance remains unknown.
To further verify this, we compare the two implementations in Tab.~\ref{tab:ablation_Rec_Inv}, excluding the MSD module to focus on the effect of the reformulation itself.
The results show that ReInversion achieves a CLIP-FG score close to that of the original Recon-Inv, indicating that the reformulation introduces only a minor and acceptable performance drop.
Meanwhile, ReInversion reduces the NFEs by half and cuts the time cost per testing case by 45\%.
This consistency and efficiency jointly validate our theoretical claim that Recon-Inv can be safely reformulated into the forward-only ReInversion process.
In addition, both methods surpass the previous powerful method FireFlow even without MSD, highlighting the effectiveness of our inversion strategy under a fair comparison setting.


\keypoint{Effect of $t_\tau$.}
The transition timestep $t_\tau$ is a crucial hyperparameter in our method, which governs the balance between preserving the source content and adopting the reference. Specifically, the diffusion process is guided towards the \textit{source} image content for time steps $t \in [0, t_\tau]$, and transitions to \textit{reference} image guidance for $t \in [t_\tau, 1]$. 
The $\tau$ values are normalized to $[0, 1]$, and the effectiveness of the editing task is highly sensitive to its setting.
As shown in Fig.~\ref{fig:ablation_tau}, a small $t_\tau$ like 0.05 triggers an early switch to reference guidance, resulting in poor preservation of source structure and producing results closer to pure reconstruction than actual editing. 
Conversely, setting $t_\tau$ too large (e.g., $t_\tau >= 0.5$) causes the model to follow the source guidance for too long. The structure becomes too rigid, making it hard for the reference to introduce modifications, and thus the final edits are barely noticeable.
Based on this analysis, the most appropriate setting for $t_\tau$ lies within the range of $[0.1, 0.3]$, which offers a good balance between source preservation and editing flexibility. 

\begin{figure}[tbp]
    \begin{center}
        \includegraphics[width=0.95\linewidth]{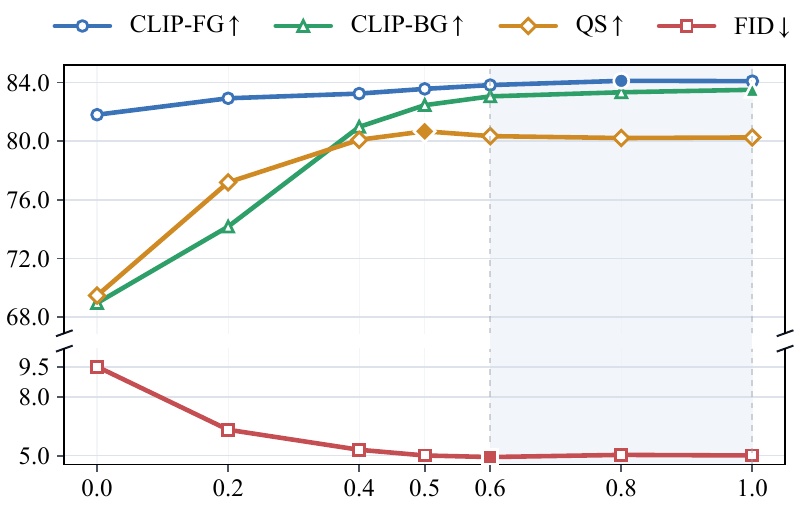}
    \end{center}
    \setlength{\abovecaptionskip}{-0.1cm}
    \caption{
        Ablation study on the effect of $\eta$. 
        $\eta$ controls the contribution of the model-predicted velocity $v_\theta$ outside the mask, interpolating between deterministic source-preserving guidance and model-driven denoising. 
        The shaded region highlights $\eta \in [0.6,1.0]$, where ReInversion achieves consistently strong performance.
    }
    \label{fig:ablation_eta}
\end{figure}

\keypoint{Effect of $\eta$.}
\label{sec:eta_effect}
We further analyze the effect of the preservation coefficient $\eta$ in MSD. 
As defined in Eq.~\ref{eq:MSD_velocity}, $\eta$ controls the interpolation outside the mask between the model-predicted velocity $v_\theta$ and the deterministic source-preserving velocity $v^*$, with larger values enforcing stronger source preservation.
Fig.~\ref{fig:ablation_eta} shows that ReInversion performs consistently well for $\eta \in [0.6,1.0]$.
A small $\eta$ leaves the unmasked region largely driven by $v_\theta$, allowing the reference condition to introduce undesired changes outside the target editing area.
In contrast, a larger $\eta$ anchors the unmasked region to the source image through $v^*$, thereby improving background preservation and overall editing consistency.

\begin{figure*}[]
    \begin{center}
        \includegraphics[width=0.98\linewidth]{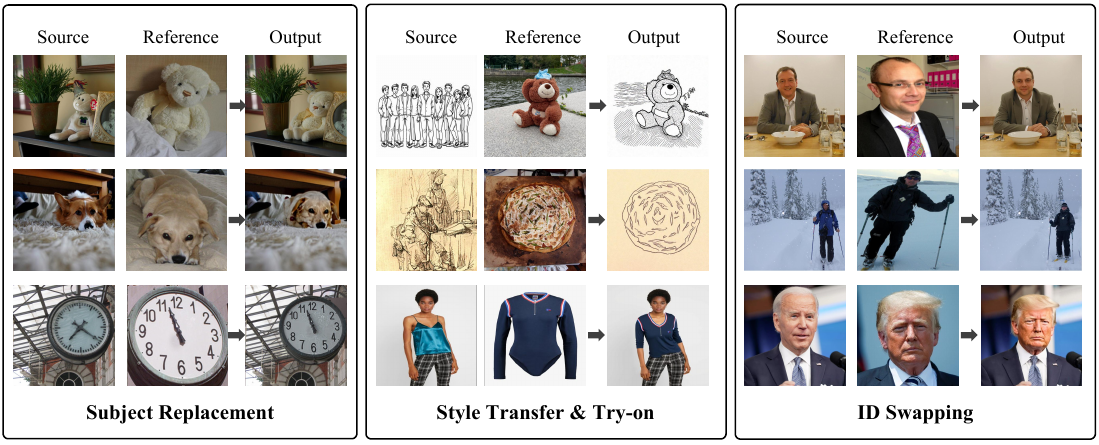}
    \end{center}
\setlength{\abovecaptionskip}{-0.1cm}
\caption{Versatile exemplar-guided editing results of ReInversion across three representative scenarios: subject replacement, style transfer and try-on, and identity transfer. Each triplet shows the source image, the reference exemplar, and the output generated by ReInversion.}
    \label{fig:versatile_results}
\end{figure*}

\subsection{Versatile Editing Scenarios}

Beyond standard object-level EIE, we further examine the applicability of ReInversion to various exemplar-guided editing scenarios. As shown in Fig.~\ref{fig:versatile_results}, the same training-free inference framework can support three representative types of visual guidance: subject replacement, style transfer and try-on, and identity transfer.

\keypoint{Subject Replacement}.
ReInversion can replace the target subject in the source image with the appearance specified by the reference exemplar while preserving the original scene layout and background. As shown in the left part of Fig.~\ref{fig:versatile_results}, the model successfully transfers different target subjects, such as plush toys and dogs, into the source context. It can also handle object-level appearance replacement, as demonstrated by the clock example, where the reference appearance is introduced while the surrounding scene remains consistent.

\keypoint{Style Transfer and Try-on}.
ReInversion can also use the reference exemplar to provide style or appearance guidance. In the middle part of Fig.~\ref{fig:versatile_results}, the model transfers visual styles from references to source contents, producing outputs that preserve the source structure while adopting the reference appearance. The try-on example further shows that ReInversion can transfer clothing appearance onto the source person without task-specific training, indicating its flexibility in appearance-level editing.

\keypoint{Identity Transfer}.
Finally, ReInversion can be applied to identity-guided editing, where the reference image provides identity-related appearance cues. As shown in the right part of Fig.~\ref{fig:versatile_results}, the outputs preserve the pose, background, and overall composition of the source images while incorporating the identity characteristics from the references. These examples suggest that ReInversion can handle high-level semantic appearance transfer beyond object replacement.

Overall, these results demonstrate that ReInversion is not limited to a single editing category. Instead, it provides a unified training-free framework that can flexibly interpret the reference exemplar as different forms of visual guidance, including subject appearance, style, clothing, and identity-related cues.

\subsection{Further Analysis}

\begin{table}[htbp]
\centering
\caption{The performance of ReInversion on different backbones (Qwen-Image-Edit~\cite{Wu2025qwen} and Flux-Kontext~\cite{Kontext}) on different inference steps (8, 18, 28).}
\label{tab:quanti_qwen}

\normalsize
\setlength{\tabcolsep}{4.3pt}
\renewcommand{\arraystretch}{1.2}

    \begin{tabular}{l|cc|cc|cc}
        \midrule
        Backbone & FID$\downarrow$ & QS$\uparrow$ & CLIP-FG$\uparrow$ & CLIP-BG$\uparrow$ & Steps \\
        \midrule
        \midrule
        \multirow{3}{*}{Qwen} & 4.88 & 78.06 & 72.56 & 83.83 & 8 \\
          & 4.84 & 76.15 & 72.47 & 83.76 & 18 \\
          & 4.65 & 78.20 & 82.30 & 83.97 & 28 \\
        \midrule
        \multirow{3}{*}{Flux} & 5.12 & 78.63 & 83.03 & 83.43 & 8 \\
          & 5.01 & 80.25 & 84.09 & 83.50 & 18 \\
          & 5.05 & 80.73 & 84.68 & 83.51 & 28 \\
        \midrule
    \end{tabular}

\end{table}

\keypoint{Effect of Backbone Architectures.}
We analyze the behavior of ReInversion across different editing backbones, including Qwen-Image-Edit and Flux-Kontext, under multiple inference budgets (8/18/28 steps). 
For Qwen-Image-Edit, we further apply 4-bit NF4 quantization to both the DiT and text encoder to reduce memory usage and enable efficient inference.
The quantitative results are reported in Tab.~\ref{tab:quanti_qwen}.
Overall, ReInversion can be readily applied to heterogeneous transformer-based editors while maintaining strong editing quality.
On Flux-Kontext, ReInversion remains effective even under very small inference budgets. 
The 8-step and 18-step settings already achieve results comparable to longer runs: increasing the number of steps from 8 to 18 improves QS from 78.63 to 80.25 and CLIP-FG from 83.03 to 84.09, while the 18-step performance is nearly identical to the 28-step setting.
This suggests that ReInversion can achieve high-quality EIE with low computational cost on Kontext.
Qwen-Image-Edit also produces competitive editing results, showing that ReInversion is not tied to a specific backbone.
However, Qwen exhibits weaker CLIP-FG under 8/18-step inference, mainly because it is not optimized for aggressive low-NFE sampling.
Its default inference schedule uses 50 steps; therefore, directly reducing the number of steps can degrade reconstruction quality and semantic fidelity.
These results indicate that ReInversion is broadly applicable, while its short-step behavior is still affected by the intrinsic sampling properties of the underlying backbone.

\begin{table}[h]
\centering
\caption{ReInversion consistently outperforms previous methods, showing that its advantage is not restricted to the curated COCOEE$^\dagger$ subset.}

\normalsize
\setlength{\tabcolsep}{4.0pt}
\renewcommand{\arraystretch}{1.2}

    \begin{tabular}{l|cc|cc}
        \toprule
        Method & FID$\downarrow$ & QS$\uparrow$ & CLIP-FG$\uparrow$ & CLIP-BG$\uparrow$ \\
        \midrule
        \midrule
        RF-Inversion & 9.13 & 56.84 & 74.40 & 64.60 \\
        FTEdit & 23.25 & 23.75 & 73.49 & 69.10 \\
        RF-Solver & 9.58 & 64.84 & 78.78 & 65.03 \\
        FireFlow & 6.82 & 64.05 & 78.07 & 66.95 \\
        \midrule
        Ours (w/o MSD) & 6.33 & 63.79 & 79.06 & 71.46 \\
        Ours (w/ MSD) & 3.77 & 74.05 & 80.27 & 83.69 \\
        \bottomrule
    \end{tabular}

\label{tab:original_cocoee}
\end{table}

\keypoint{Evaluation on Original COCOEE.}
In the main experiments, we evaluate ReInversion on COCOEE$^\dagger$, a curated subset of COCOEE where low-quality and ambiguous reference cases are removed to ensure reliable evaluation.
To further verify that the observed improvements are not caused by this filtering process, we additionally report quantitative results on the original COCOEE benchmark in Tab.~\ref{tab:original_cocoee}.
As shown in the table, ReInversion remains highly effective on the original benchmark.
Without MSD, our method already achieves competitive performance, obtaining the best FID of 6.33 and the highest QS of 63.79 among all compared methods.
After incorporating MSD, ReInversion further improves all metrics, reducing FID from 6.33 to 3.77 and increasing QS from 63.79 to 74.05.
It also achieves the strongest foreground and background consistency, with CLIP-FG of 80.27 and CLIP-BG of 83.69.
These results demonstrate that ReInversion maintains strong performance even under the more challenging and less curated setting of the original COCOEE benchmark, and its advantages over prior inversion-based and flow-based editing methods are not limited to the curated evaluation subset.
The detailed filtering list and additional dataset information are provided in our code repository.

\section{Conclusion}


In this work, we present ReInversion, a training-free framework for effective and efficient exemplar-guided image editing (EIE).
First, we introduce Reconstruction-Based Inversion (Recon-Inv) to mitigate trajectory drift during inversion.
Second, we reformulate Recon-Inv into ReInversion, enabling high-quality EIE with only 14 denoising steps.
Third, we propose Mask-Guided Selective Denoising (MSD) to preserve structural and color consistency in unedited regions.
Extensive experiments demonstrate that ReInversion achieves superior editing quality, structural consistency, and computational efficiency. 
Notably, it generalizes well across different diffusion backbones and remains effective under varying denoising steps, even with a very small number of denoising steps.
These results demonstrate that training-free methods can achieve both high-quality and efficient EIE.


\bibliography{reference}
\bibliographystyle{IEEEtran}

\end{document}